\documentclass[letterpaper, 10 pt, conference]{ieeeconf}  

\IEEEoverridecommandlockouts                              

\usepackage{graphicx}      
\usepackage{algorithm} 
\usepackage{algpseudocode} 
\usepackage{varwidth} 
\usepackage{amsfonts}
\usepackage{amsmath}
\usepackage{url}

\usepackage[caption=false,font=footnotesize]{subfig}

\graphicspath{{figures/}}
\title{\LARGE \bf
	Real-Time Grasp Planning for Multi-Fingered Hands by Finger Splitting
}

\author{Yongxiang Fan, Te Tang, Hsien-Chung Lin, Masayoshi Tomizuka 
	\thanks{Yongxiang Fan, Te Tang, Hsien-Chung Lin, and Masayoshi Tomizuka are with Department of Mechanical Engineering, 
		University of California, Berkeley, Berkeley, CA 94720, USA
		{\tt\small {yongxiang\_fan, tetang, hclin, tomizuka}@berkeley.edu}}%
}

\begin{document}
	\maketitle
	\thispagestyle{empty}
	\pagestyle{empty}
	
	\begin{abstract}
		Grasp planning for multi-fingered hands is computationally expensive due to the joint-contact coupling, surface nonlinearities and high dimensionality, thus is generally not affordable for real-time implementations. Traditional planning methods by optimization, sampling or learning work well in planning for parallel grippers but remain challenging for multi-fingered hands. This paper proposes a strategy called finger splitting, to plan precision grasps for multi-fingered hands starting from optimal parallel grasps. The finger splitting is optimized by a dual-stage iterative optimization including a contact point optimization (CPO) and a palm pose optimization (PPO), to gradually split fingers and adjust both the contact points and the palm pose. The dual-stage optimization is able to consider both the object grasp quality and hand manipulability, address the nonlinearities and coupling, and achieve efficient convergence within one second. Simulation results demonstrate the effectiveness of the proposed approach. The simulation video is available at~\cite{youtube}. 
	\end{abstract}
	
	\section{INTRODUCTION}
	Grasp planning for multi-fingered hands is important for robotic grasping and manipulation in order to increase the dexterity and collaborate with humans. Current grasping in factories usually requires considerable time to design specific grippers and program motion sequences for assembly or pick-and-place tasks. A general purposed multi-fingered hand will greatly simplify the procedure and improve the adaptability to new tasks. However, the grasp planning for general purposed multi-fingered hands is challenging due to the large variations of objects, coupling between the hand and objects, and high dimensionality of the hand-object system. 
	More specifically, a multi-fingered hand for general purposes should be able to grasp different objects with various surfaces and sizes, from simple boxes to toys or tools with complicated shapes. Moreover, a stable grasp relies on proper contacts between the fingers and the object. The contacts that associate the hand and the  object has to be on the object surface and also reachable by all the fingers attached to the palm. Considering the configuration of multiple fingers and aforementioned constraints, the searching for optimal grasp in the sense of maximizing the object grasp quality and the hand manipulability, becomes a high-dimensional problem, thus has strong limitation for real-time applications. 
	
	Several different approaches have been proposed to speed up the searching for optimal grasps. 
	In~\cite{hang2016hierarchical}, a hierarchical finger space is generated on object surface and the grasps are synthesized by a multi-level refinement strategy. The mapping from the contact pairs to hand configuration is obtained by an object-specific reachability table generated offline. 
	A part-based grasp planning method is proposed in~\cite{aleotti2011part} using Reeb graph, with the assumption that the grasping is constrained in single part of the object. 
	Object are represented by inscribing spheres in~\cite{przybylski2011planning} and the grasp searching is conducted on qualified subset of spheres.  
	The idea of eigengrasps is described in~\cite{ciocarlie2007dexterous} to simplify the joint space searching, and the grasp is optimized by solving a nonlinear optimization with over one hundred seconds.  In~\cite{Shi2017Real}, the power grasps are planned by sequentially searching the palm pose and contact positions. Instead of optimizing for grasp qualities, several heuristics such as the intersected volume and the finger curling planes are utilized, and the contact point searching assumes planar finger motion and fixed palm locations. 
	
	Compared with  multi-fingered hands, the grasp planning for parallel grippers 
	is conducted in much lower dimensional space due to the simpler gripper structures and fewer constraints. 
	In~\cite{levine2016learning}, a deep reinforcement learning approach is proposed to directly learn the grasping policy from images. The grasping skills are learned from exploration measured by empirical success rate. Some others utilize databases to learn grasps for similar objects, these include the Columbia grasp database~\cite{goldfeder2009columbia} and dexterity network (Dex-Net)~\cite{mahler2016dex}. 
	
	In this paper, we will study the knowledge transfer from grasps for parallel grippers to those for multi-fingered hands. We propose a strategy called finger splitting, to generate precision grasps for a  multi-fingered hand from parallel ones.  
	To be more specific, the multi-fingered hand is initialized by a parallel grasp with two contacts by assuming that the fingers are separated into two groups around contacts. The grasps for parallel grippers can be computed from database for parallel grippers~\cite{mahler2016dex} or planned by iterative surface fitting (ISF)~\cite{Fan2018Grasp}.
	Then the splitting algorithm gradually spreads all fingers from the original two contacts by optimizing both the object grasp quality and the hand manipulability. An optimal grasp with proper palm pose, joint angles and contacts will be generated after splitting.

	The contributions of this paper are as follows. First, a novel finger splitting strategy is proposed to transfer the knowledge from grasp databases for parallel grippers to planning of precision grasps for multi-fingered hands, where only fingertips are contacted with the object. The transferring leads to better feasibility guarantee and faster convergence. Moreover, a dual-stage iterative optimization algorithm is proposed to control the splitting behavior and search for optimal configurations to maximize both the object grasp quality and the hand manipulability. The dual-stage optimization effectively reduces the coupling between the palm and contacts and decomposes the high-dimensional optimization problem into two lower dimensional optimizations with less constraints. Furthermore, the decomposed optimizations are solved by customized gradient projections. The customization avoids modeling of the object surface and is computationally efficient. The average computation time is less than one second for the objects of different categories, and is appealing for real-time applications. The proposed approach is verified by simulations and the simulation video is available at~\cite{youtube}. 
	
	The remainder of this paper is described as follows. First, the problem of a general grasp planning for multi-fingered hands is stated in Section~\ref{sec:problem_statement}, followed by a detailed explanation of the proposed finger splitting in Section~\ref{sec:finger_splitting}. Simulation results are presented in Section~\ref{res:sim_exp}. Section~\ref{sec:conclusion} concludes this paper and proposes future works.

	\section{PROBLEM STATEMENT}
	\label{sec:problem_statement}
\begin{figure}[htb]
		\begin{center}
			\includegraphics[width=1.2in]{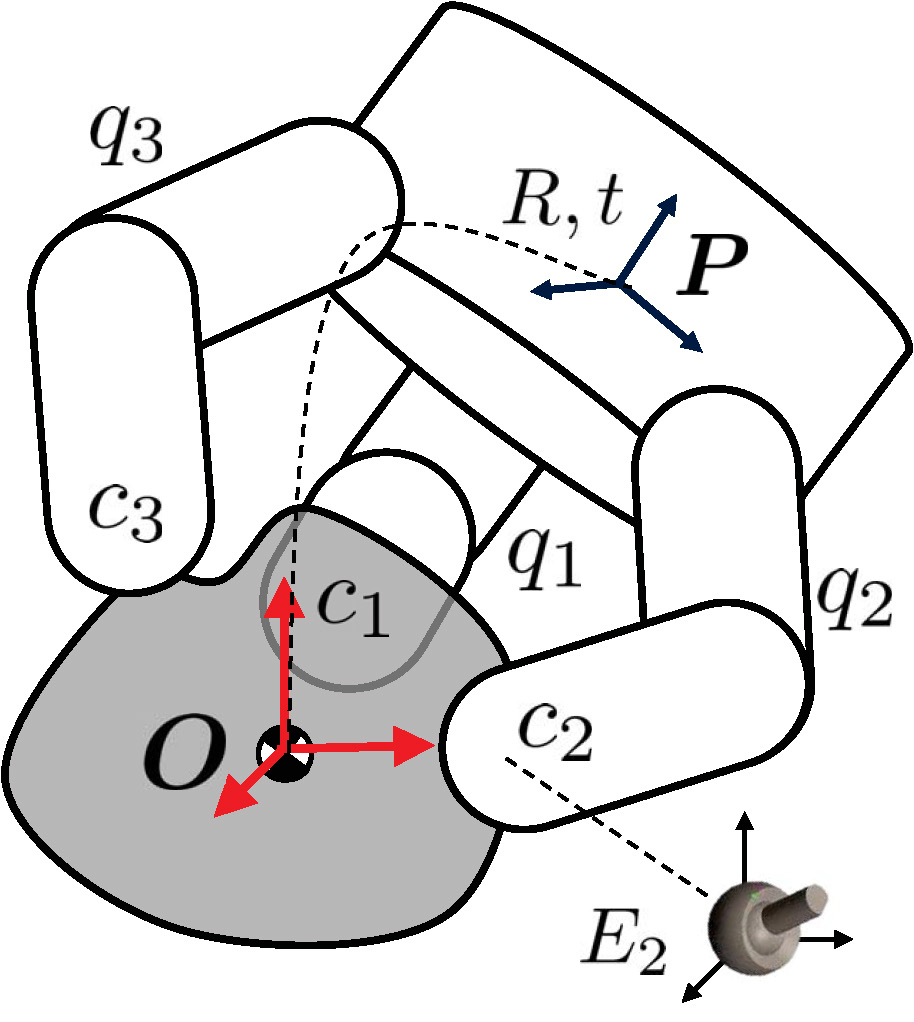}
			\caption{Illustration of grasp planning problem using a three-fingered hand. }
			\label{fig:object_hand}
		\end{center}
	\end{figure}

	This section describes a general precision grasp planning problem with a three-fingered robotic hand illustrated in Fig.~\ref{fig:object_hand}. The goal of the planning is to search for optimal contacts on the object and the associated hand configuration that maximize the object grasp quality and the hand manipulability. The contact position vector is denoted by $\boldsymbol{c} = \left[c_1^T, c_2^T, c_3^T\right]^T$, where $c_i\in \mathbb{R}^3$ is the contact for the $i$-th finger. The joint angle vector of the hand is denoted by $\boldsymbol{q} = \left[q_1^T, q_2^T, q_3^T\right]^T$, where $q_i\in\mathbb{R}^{N_{\text{jnt},i}}$ is the joint angle for the $i$-th finger and $N_{\text{jnt},i}$ is the number of joints for the $i$-th finger. The pose of the palm is represented by rotation $R\in SO(3)$ and translation $t\in \mathbb{R}^3$. 
	The contacts can be modeled as point contact with friction model~\cite{murray1994mathematical} and would exhibit rotational freedoms. They can be represented as ball joints and characterized by Euler angles $\boldsymbol{E} = \left[E_1^T, E_2^T, E_3^T\right]^T$, where $E_i\in \mathbb{R}^3$ is the Euler angle for the $i$-th contact. 
	Mathematically, the grasp planning problem can be formulated as: 
	\begin{subequations}
		\label{eq:general_form}
		\begin{align}
		\max_{R, t, \boldsymbol{q}, \boldsymbol{E},\boldsymbol{c}} &\  Q(\boldsymbol{c}, \boldsymbol{q}) \label{eq:general_cost}\\
		s.t. \quad 
		& (R,t) = FK_{c2p}(\boldsymbol{q}, \boldsymbol{E}, \boldsymbol{c}) \label{eq:general_FK}\\
		& c_i \in \partial O \label{eq:general_surface} \quad i = 1 \cdots 3\\
		& q_i \in [q_{\text{min},i}, q_{\text{max},i}] \label{eq:general_limit} \quad i = 1\cdots 3
		\end{align}
	\end{subequations}
	where $Q(\boldsymbol{c},\boldsymbol{q})$ is the overall grasp quality containing both the object grasp quality and the hand manipulability,  $FK_{c2p}$ is the forward kinematics from contacts to the palm, and $q_{\text{min},i}, q_{\text{max},i}\in~\mathbb{R}^{N_{\text{jnt},i}}$ are joint limits for the $i$-th finger. Constraint~(\ref{eq:general_FK}) connects the palm and contacts through finger joints, (\ref{eq:general_surface}) constrains contacts on the object surface $\partial O$, and (\ref{eq:general_limit}) shows the joint limits.
	With the palm pose, joints and contacts as optimization variables, the forward kinematics~(\ref{eq:general_FK}) and complicated object surface~(\ref{eq:general_surface}) as constraints, the problem~(\ref{eq:general_form}) becomes a high-dimensional nonlinear programming. The optimization may become more challenging when considering collision (i.e. unexpected contact) between the hand and object.
	
	\begin{figure}[t]
		\begin{center}
			\includegraphics[width=3in]{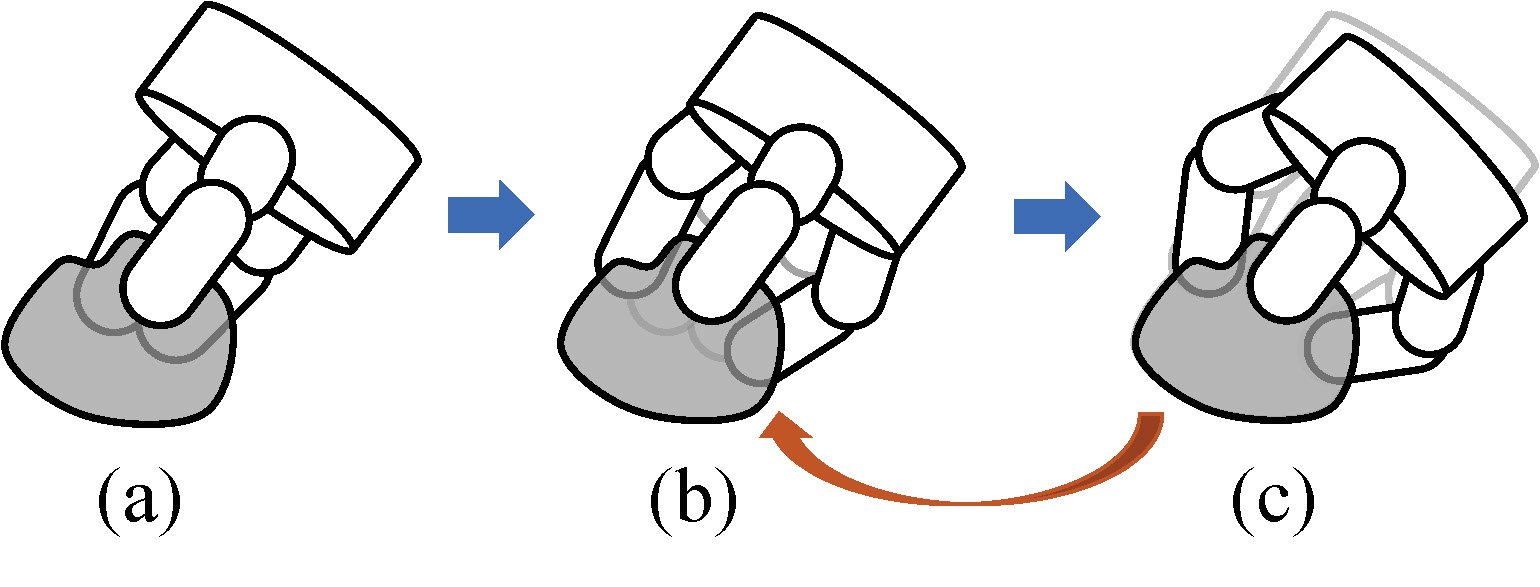}
			\caption{Finger splitting using dual-stage iterative optimization. The finger splitting is initialized by a parallel grasp shown in~(a) and is optimized by iteratively executing a contact point optimization shown in (b) and a palm pose optimization shown in~(c).}
			\label{fig:dual_stage}
		\end{center}
	\end{figure}
	
	In this paper, we propose a finger splitting strategy to solve~(\ref{eq:general_form}). The finger splitting separates all the fingers into two groups and gradually spread them to maximize the object grasp quality and the hand manipulability. The idea of finger splitting is shown in Fig.~\ref{fig:dual_stage}.  
	The initial parallel grasps can be generated by database (e.g. Dex-Net~\cite{mahler2016dex}) or computed by ISF~\cite{Fan2018Grasp}, as shown in Fig.~\ref{fig:dual_stage}(a), after which a dual-stage iterative optimization is introduced to search for new contacts and hand configuration. Stage one is named as contact point optimization (CPO). The objective of the CPO is to maximize the object grasp quality and hand manipulability by searching for contacts $\boldsymbol{c}$ and joints $\boldsymbol{q}$ while keeping palm pose $R,t$ fixed, as shown in Fig.~\ref{fig:dual_stage}(b). Stage two is called palm pose optimization (PPO). The objective of the PPO is to improve the hand manipulability by searching over the palm pose $R,t$ and joints $\boldsymbol{q}$ while keeping the contacts $\boldsymbol{c}$ fixed, as shown in Fig.~\ref{fig:dual_stage}(c). The CPO and PPO will be iteratively executed until converge or termination conditions are reached.

	\section{FINGER SPLITTING}
	\label{sec:finger_splitting}
	
	\subsection{Contact Point Optimization (CPO)}
	The CPO searches for desired contacts and joints to maximize the object grasp quality and the hand manipulability by assuming that the palm pose $R,t$ is static. 
	The CPO is formulated as: 
	\begin{subequations}
		\label{eq:cpo_nlp}
		\begin{align}
		\max_{\boldsymbol{c}, \boldsymbol{q}} &\  Q(\boldsymbol{c}, \boldsymbol{q}) \label{eq:cpo_cost}\\
		s.t. \quad 
		& \boldsymbol{c}= FK_{c2p}(\boldsymbol{q}, R_0, t_0) \label{eq:cpo_fk}\\
		& c_i \in \partial O \quad i = 1 \cdots 3 \label{eq:cpo_surface}\\
		& q_i \in [q_{\text{min},i}, q_{\text{max},i}] \quad i = 1 \cdots 3 \label{eq:cpo_limit}
		\end{align}
	\end{subequations}
	where $Q(\boldsymbol{c},\boldsymbol{q}) = w_1Q_o(\boldsymbol{c}) + w_2Q_h(\boldsymbol{q})$ indicates the overall grasp quality composed by the object grasp quality $Q_o$ and the hand manipulability $Q_h$. The object grasp quality $Q_o$ can be represented by the triangle area formed by the contacts: $Q_o(\boldsymbol{c}) = 2\text{Area}\left(\{c_i\}_{i=1\cdots 3}\right)$~\cite{supuk2005estimation}, and the hand manipulability $Q_h$ can be represented by the deviation from the center of the joints: $Q_h(\boldsymbol{q}) = -0.5\sum_{i = 1}^{3}\sum_{j = 1}^{N_{\text{jnt},i}}\left((q_i^j -\bar{q}_i^j)/(q_{\text{max},i}^j - q_{\text{min},i}^j )\right)^2$ based on~\cite{liegeois1977automatic}, where $q_i^j$ is the $j$-th joint angle of the $i$-th finger, $q_{\text{min},i}^j$ and $ q_{\text{max},i}^j$ are the limits of $q_i^j$, $\bar{q}_i^j = (q_{\text{max},i}^j + q_{\text{min},i}^j)/2$ is the middle position of the corresponding joint. 
	$FK_{q2c}$ is the forward kinematics from joint $\boldsymbol{q}$ to contact $\boldsymbol{c}$, and $R_0,t_0$ denote the fixed rotation and translation of the palm.
	
	Optimization~(\ref{eq:cpo_nlp}) remains a nonlinear programming due to the nonlinearities of~(\ref{eq:cpo_fk}) and~(\ref{eq:cpo_surface}). Moreover, the object surface has to be fitted and fed into the optimization before running the gradient based method. In~\cite{fan2017real, fan2017real2}, we introduced a velocity-level finger gaits planner for dexterous manipulation. A similar approach is proposed in this paper to solve~(\ref{eq:cpo_nlp}). The idea is to iteratively search on tangent space and project to the nonlinear constraints.
	\subsubsection{Tangent space searching}
	The tangent space searching is to find the displacement vectors in order to maximize the quality in the next time step. More specifically, 
	\begin{subequations}
		\label{eq:cpo_lp}
		\begin{align}
		\max_{{\boldsymbol{d_c}}, {\boldsymbol{d_q}}} &\  \nabla_{\boldsymbol{c}} {Q}({\boldsymbol{c}}, {\boldsymbol{q}})\boldsymbol{d_c} + \nabla_{\boldsymbol{q}}Q({\boldsymbol{c}}, {\boldsymbol{q}})\boldsymbol{d_q} \label{eq:cpo_lpcost}\\
		s.t. \quad 
		& {\boldsymbol{d_c}} = J_{q2c}(\boldsymbol{q}, R_0, t_0){\boldsymbol{d_q}} \label{eq:cpo_lpfk}\\
		& \boldsymbol{n}^T(\boldsymbol{c}){\boldsymbol{d_c}} = 0 \label{eq:cpo_lpsurface}\\
		& \|\boldsymbol{d}\| \leq \sigma_{cpo} \label{eq:cpo_lpstep}
		\end{align}
	\end{subequations}
	where $\boldsymbol{d_c} = \boldsymbol{c}(t + T_s) - \boldsymbol{c}(t)$ and $\boldsymbol{d_q} = \boldsymbol{q}(t + T_s) - \boldsymbol{q}(t)$ are displacement vectors for contacts and joints, $T_s$ is the simulation time step, and $\boldsymbol{d} = [\boldsymbol{d_c}^T, \boldsymbol{d_q}^T]^T$. $J_{q2c} = \text{diag}\left(\left[J_{q2c,1},J_{q2c,2},J_{q2c,3}\right]\right)$ denotes a geometric Jacobian from $\boldsymbol{q}$ to $\boldsymbol{c}$, and $J_{q2c,i}\in \mathbb{R}^{3\times N_{\text{jnt},i}}$ is the translational Jacobian for the $i$-th finger.  $\boldsymbol{n}(\boldsymbol{c}) = \text{diag}\left(\left[\nabla_{c_1}^T (\partial O), \nabla_{c_2}^T (\partial O), \nabla_{c_3}^T (\partial O)\right]\right)$ is the surface normal matrix of the contacts, and $\nabla_{c_i} (\partial O) \in \mathbb{R}^{1\times3}$ is the normalized surface gradient w.r.t. $c_i$. $\sigma_{cpo}$ is step size constraint. 
	
	
	Optimization~(\ref{eq:cpo_lp}) is the tangent approximation of~(\ref{eq:cpo_nlp}) and can be solved analytically~\cite{luenberger1984linear} by:  
	\begin{equation}
	\label{eq:cpo_solve}
	\begin{aligned}
	& \boldsymbol{d^*} = \alpha^*\boldsymbol{d_0}\\
	& \boldsymbol{d_0} = (I - A^T (AA^T)^{-1}A)\nabla_{\boldsymbol{x}}Q^T(\boldsymbol{x}) \\
	\end{aligned}
	\end{equation}
	where $\boldsymbol{d^*} = [{\boldsymbol{d_c^*}}^T; {\boldsymbol{d_q^*}}^T]^T$ is the optimal tangent displacement vector and $A = [n^T(\boldsymbol{c}), 0; -I_9,  J_{q2c}]$, $\boldsymbol{x} = [\boldsymbol{c}^T, \boldsymbol{q}^T]^T$. $\alpha^*$ is the optimal step size obtained by: 
	\begin{equation*}
	\begin{aligned}
	\alpha^* = 
	\begin{cases}
	\sigma_{cpo}/\|\boldsymbol{d_0}\|, & \text{if } F_{ls} = 0\\
	\underset{0 \leq \alpha \leq \sigma_{cpo}/\|\boldsymbol{d_0}\|}{\arg\max} \left(Q\left(Proj\left(\boldsymbol{x} + \alpha \boldsymbol{d_0}\right)\right)\right), & \text{if } F_{ls} = 1
	\end{cases}
	\end{aligned}
	\end{equation*}
	where $F_{ls}$ is the option for line search, and the $Proj$ represents projection operation introduced below. The line search can be disabled (i.e. $F_{ls} = 0$) empirically to accelerate the computation, though it would introduce certain oscillation. 
	
	\subsubsection{Nonlinear constraints projection}
	Several steps are taken in order to project the tangent state $\boldsymbol{x} + \alpha \boldsymbol{d_0}$ back to the nonlinear constraints $h(\boldsymbol{c,q}) = \{{\boldsymbol{c} \in \partial O, \boldsymbol{c} = FK_{q2c}(\boldsymbol{q})}\}$. 
	First, the reference nearest neighbor $\boldsymbol{c_\text{ref}}$  of the optimal tangent contacts $\boldsymbol{c}+\boldsymbol{d_c^*}$ is searched by KD-Tree on the object surface. Second, the found nearest neighbor $\boldsymbol{c}_{\text{ref}}\in \partial O$ is tracked by a simple stiffness controller $\dot{\boldsymbol{q}}_{\text{des}} = J_{q2c}^{-1}(\boldsymbol{c})K_{cpo}(\boldsymbol{c}_{\text{ref}} - \boldsymbol{f})$, where $\boldsymbol{f}$ is the current fingertip position vector. Last, the joint state in simulator is updated by $\boldsymbol{q_\text{des}} \leftarrow \boldsymbol{q} + \dot{\boldsymbol{q}}_{\text{des}}T_s$ if the termination condition is not satisfied. 
	
	\subsubsection{Collision detection}
	Unexpected contact between the object and the hand is called collision in this paper since it might effect the execution of the optimized grasps. A simple collision detection algorithm is introduced to stop the finger splitting upon collision. The algorithm takes supervoxel representation of the object~\cite{Papon13CVPR} as input, and check the inclusion relation of each supervoxel in finger links. The supervoxel representation usually only includes hundreds of points and the finger links are approximated by  boxes or cylinders, thus the collision detection can be extremely fast. The collision detection is implemented in both CPO and PPO. 
	
\begin{figure}[t]
		\begin{center}
			\includegraphics[width=2.5in]{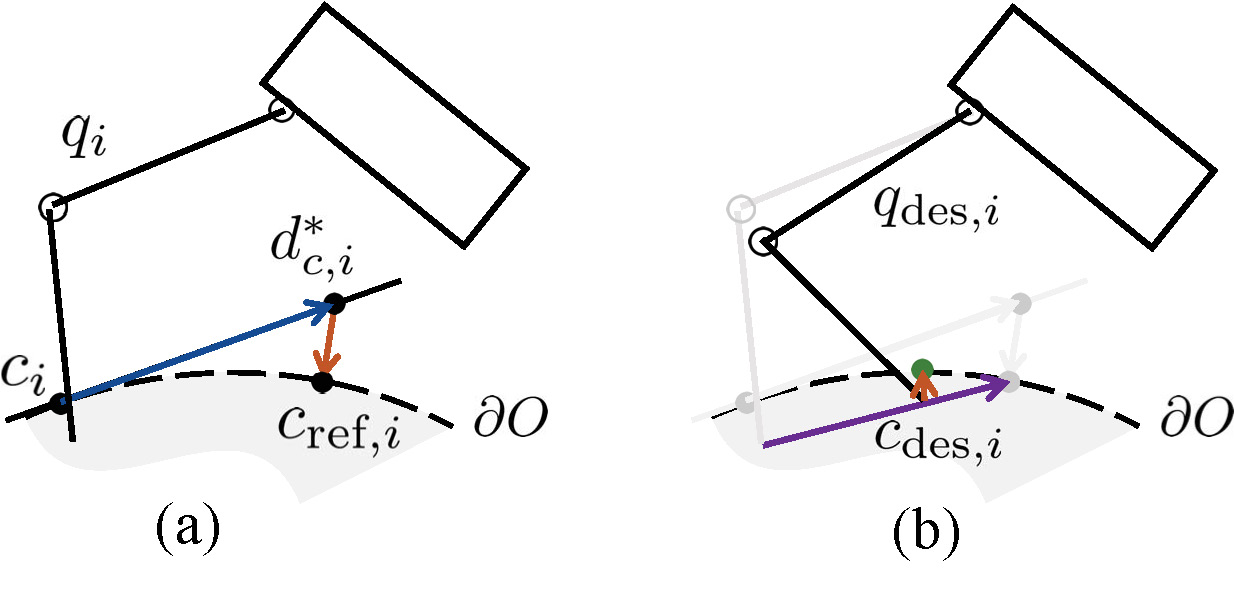}
			\caption{Illustration of the CPO algorithm. (a) shows tangent space searching and reference generation, and (b) shows reference tracking and contact estimation.}  %
			\label{fig:cpo_illustration}
		\end{center}
	\end{figure}

	
	Finally, the overall CPO algorithm is summarized in Alg.~(\ref{alg:cpo}). The CPO will be terminated once the quality increment is less or equal than $\delta_{c}$, or the angle between the surface normal $\boldsymbol{n_\text{des}}$  and a predefined artificial fingertip normal $\boldsymbol{n_f}(\boldsymbol{q_\text{des}})$ is larger than $\gamma$, as shown in Line~(\ref{cpo:t}).
	A graphical illustration of the CPO algorithm is shown in Fig.~\ref{fig:cpo_illustration}. Figure~\ref{fig:cpo_illustration}(a) describes the tangent space searching (blue arrow line) and the nearest neighbor search (red arrow line) of the reference contact corresponding to Line~(\ref{cpo:tangent}-\ref{cpo:reference}), and Fig.~\ref{fig:cpo_illustration}(b) shows the tracking of the reference contact (purple arrow line) and estimation of the desired contact (red arrow line) corresponding to Line~(\ref{cpo:desire1}-\ref{cpo:t}). The CPO is able to handle the case where the initial fingertip positions are not on the object surface by using the tracking control. 
	\begin{algorithm}
		\caption{Contact Point Optimization (CPO)}\label{alg:cpo}
		\begin{algorithmic}[1]
			\State \textbf{Input: } Static palm pose $R_0, t_0$, initial states $\boldsymbol{c}, \boldsymbol{q}$\label{cpo:input}
			\While {$it_{cpo}\texttt{++} < M$}
			\State Search tangent motion $\boldsymbol{d_c^*}$ in~(\ref{eq:cpo_lp}) by solving~(\ref{eq:cpo_solve}) \label{cpo:tangent}
			\State Search reference contact $\boldsymbol{c_{\text{ref}}} \leftarrow NN_{\partial O}(\boldsymbol{c} + \boldsymbol{d_c^*})$  \label{cpo:reference}
			\State Track reference contact ${\boldsymbol{\dot{q}_{\text{des}}}} \leftarrow J_{q2c}^{-1}K_{cpo}(\boldsymbol{c_{\text{ref}}}  - \boldsymbol{f})$ 
			\State Compute desired state $\boldsymbol{q_{\text{des}}},  \boldsymbol{c_{\text{des}}}, \boldsymbol{n_{\text{des}}}$ by: \label{cpo:desire1} \par
			\hskip\algorithmicindent$ {\boldsymbol{q_{\text{des}}}} \leftarrow {\boldsymbol{\dot{q}_{\text{des}}}} T_s + \boldsymbol{q}$ \par
			\hskip\algorithmicindent$ (\boldsymbol{c_{\text{des}}}, \boldsymbol{n_{\text{des}}}) \leftarrow NN_{\partial O}(FK_{q2c}(\boldsymbol{q_\text{des}}, R_0, t_0))$
			\State Test termination:  $\Delta Q_\text{des} = Q_\text{des}(\boldsymbol{c_\text{des}}, \boldsymbol{q_\text{des}}) - Q(\boldsymbol{c},\boldsymbol{q}) $ \par
			\hskip \algorithmicindent$stop \leftarrow (\Delta Q_\text{des} \leq \delta_c) \ \|\  (|\boldsymbol{n_\text{des}}^T\boldsymbol{n_f}(\boldsymbol{q_\text{des}})| \geq \gamma)$ \par
			\hskip\algorithmicindent $collide = col\_detect(\boldsymbol{q_\text{des}},R_0, t_0)$ \par
			\label{cpo:t}
			\If{$stop\  \|\  collide $}   \label{cpo:t1}
			\textbf{break} \EndIf  \label{cpo:t2}
			\State	Set state $\boldsymbol{q}\leftarrow \boldsymbol{q_\text{des}}, \  \boldsymbol{c}\leftarrow \boldsymbol{c_\text{des}}, \  \boldsymbol{n} \leftarrow \boldsymbol{n_\text{des}}$ \label{cpo:t3}
			\EndWhile
		\end{algorithmic}
	\end{algorithm}

	\subsection{Palm Pose Optimization (PPO)}
	The CPO optimizes contact points by assuming that the palm pose is constant. However, the palm might not be in the best pose, thus the CPO can only find a local optimum in subspace $\mathcal{S} = \{\boldsymbol{c}\in \partial O,\boldsymbol{q} \in [\boldsymbol{q_\text{min}}, \boldsymbol{q_\text{max}}]\ |\ R_0, t_0 \}$. In this section, the palm pose optimization will be introduced to maximize the overall grasp quality assuming that the contact points are static (i.e. object grasp quality $Q_o$ is constant). The PPO can be formulated as: 
	\begin{subequations}
		\label{eq:ppo_nlp}
		\begin{align}
		\max_{R\in SO(3), t, \boldsymbol{q}, \boldsymbol{E}} &\  Q(\boldsymbol{c}_0, \boldsymbol{q}) \label{eq:ppo_cost}\\
		s.t. \quad 
		& (R,t) = FK_{c2p}(\boldsymbol{c}_0,\boldsymbol{q}, \boldsymbol{E}) \label{eq:ppo_fk}\\
		& q_i \in [q_{\text{min},i}, q_{\text{max},i}]  \quad i = 1...3 \label{eq:ppo_limit}
		\end{align}
	\end{subequations}
	where $\boldsymbol{c}_0$ is the current static contacts on the object. The optimization~(\ref{eq:ppo_nlp}) is nonlinear because of the constraint~(\ref{eq:ppo_fk}) and $R\in SO(3)$.  
	Similar to CPO, the PPO in this section is solved by linearization and projection. 
	
	\subsubsection{Tangent space searching}
	Considering the relativity between the object and the hand motion, the hand palm is assumed to be static and object pose is optimized in PPO, since it is easier to implement and would increase the robustness to numerical errors. 
	Rather than taking derivative of~(\ref{eq:ppo_fk}), we notice that the linearization of~(\ref{eq:ppo_fk}) is closely related to the fundamental grasping constraint~\cite{murray1994mathematical}: 
	\begin{equation}
	\label{eq:fundamental}
	G(x_{po}, \boldsymbol{q})^TV_{po}^b = J_h(x_{po}, \boldsymbol{q})\dot{\boldsymbol{q}}
	\end{equation}
	where $x_{po} = [t_{po}^T, E_{po}^T]^T\in \mathbb{R}^{6}$ is a local parameterization of the object pose in palm frame~$\boldsymbol{P}$, with $t_{po}$ and $E_{po}$ denoting the translation  and orientation components, respectively. $V_{po}^b = [{v_{po}^b}^T, {\omega_{po}^b}^T]^T \in \mathbb{R}^{6}$ is the body velocity of the object, with $v_{po}^b$ and $\omega_{po}^b$ denoting the translational and rotational velocities, respectively. 
	$G(x_{po}, \boldsymbol{q}) \in \mathbb{R}^{6\times 9}$ and $J_h(x_{po}, \boldsymbol{q}) \in \mathbb{R}^{9\times N_\text{jnt}}$ represent grasp map and hand Jacobian~\cite{murray1994mathematical}. Equation~(\ref{eq:fundamental}) connects joint velocity and object velocity under the static contact condition. 
	The optimization~(\ref{eq:ppo_nlp}) can be solved by utilizing this connection and searching on the tangent space. If $V_{po}^b$ and $\dot{\boldsymbol{q}}$ are treated as the tangent displacements in one time step, then the tangent space searching of~(\ref{eq:ppo_nlp}) becomes:
	\begin{subequations}
		\label{eq:ppo_lp}
		\begin{align}
		\max_{V_{po}^b, {\dot{\boldsymbol{q}}}} &\  \nabla_{\boldsymbol{q}}Q({\boldsymbol{c_0}}, {\boldsymbol{q}})\dot{\boldsymbol{q}} \label{eq:ppo_lpcost}\\
		s.t. \quad 
		& G(x_{po}, \boldsymbol{q})^TV_{po}^b = J_h(x_{po}, \boldsymbol{q})\dot{\boldsymbol{q}}\\
		& \| {\dot{\boldsymbol{q}}}\| \leq \sigma_{ppo} \label{eq:ppo_lpstep}
		\end{align}
	\end{subequations}
	where $\sigma_{ppo}$ is a trust region of the joint motion. 
	Similar to~(\ref{eq:cpo_lp}), (\ref{eq:ppo_lp}) can be solved analytically as: 
	\begin{equation}
	\label{eq:ppo_solve}
	\begin{aligned}
	&  V_{po,\text{des}}^b = \sigma_{ppo}\boldsymbol{d_c}/\|\boldsymbol{d_q}\|\\
	&  \boldsymbol{d_c} = G\left(G^TG + J_hJ_h^T\right)^{-1}J_h\nabla_{\boldsymbol{q}}^TQ \\
	& \boldsymbol{d_q} = \nabla_{\boldsymbol{q}}^TQ - J_h^T\left(G^TG + J_hJ_h^T\right)^{-1}J_h\nabla_{\boldsymbol{q}}^TQ \\
	\end{aligned}
	\end{equation}

	%
	
	The desired tangent displacement of the object is represented as ${V}_{po,\text{des}}^bT_s$.
	
	\subsubsection{Nonlinear projection}
	With the desired object tangent displacement in one step ${V}_{po,\text{des}}^bT_s$, the desired object pose $g_{po,\text{des}} = (R_{po,\text{des}}, t_{po,\text{des}})$ can be obtained by 
	\begin{equation}
	\label{eq:ppo_object_update}
	\begin{aligned}
	& g_{po,\text{des}} =g_{po}e^{\hat{V}_{po,\text{des}}^bT_s}
	\end{aligned}
	\end{equation}
	where $\hat{V}_{po,\text{des}}^b \in se(3)$ is the matrix representation of $V_{po,\text{des}}^b$ shown in~\cite{murray1994mathematical} and $g_{po}\in SE(3)$ denotes the current object pose in palm frame.  
	The joint $\boldsymbol{q}_\text{des}$ is computed as follows. First, the position $\boldsymbol{c_\text{des}^p}$ and translational velocity $\boldsymbol{v_{c,\text{des}}^p}$ of the static contact in palm frame is obtained by the desired object motion $V_{po,\text{des}}^b$, after which a tracking controller is applied to compute the projected joint velocity  $\dot{\boldsymbol{q}}_\text{des}$ by tracking both the position and velocity of the reference contact:
	\begin{equation}
	\label{eq:ppo_track}
	\dot{\boldsymbol{q}}_\text{des} = J_{q2c}^{-1}\left(\boldsymbol{v_{c, \text{des}}^p} + K_{ppo}(\boldsymbol{c_\text{des}^p} - \boldsymbol{f}^p)\right)
	\end{equation}
	where  $\boldsymbol{f^p} \in \mathbb{R}^9$ is the current fingertip position vector in palm frame, and $K_{ppo}$ is the tracking gain used to reduce the misalignment between  $\boldsymbol{c^p}$ and $\boldsymbol{f^p}$ during projection. With the projected displacement $\dot{\boldsymbol{q}}_\text{des}T_s$,
	the desired finger joints can be computed as ${\boldsymbol{q}_\text{des}} = {\boldsymbol{q}} + \dot{\boldsymbol{q}}_\text{des}T_s$. 
	
\begin{figure}[t]
		\begin{center}
			\includegraphics[width=3in]{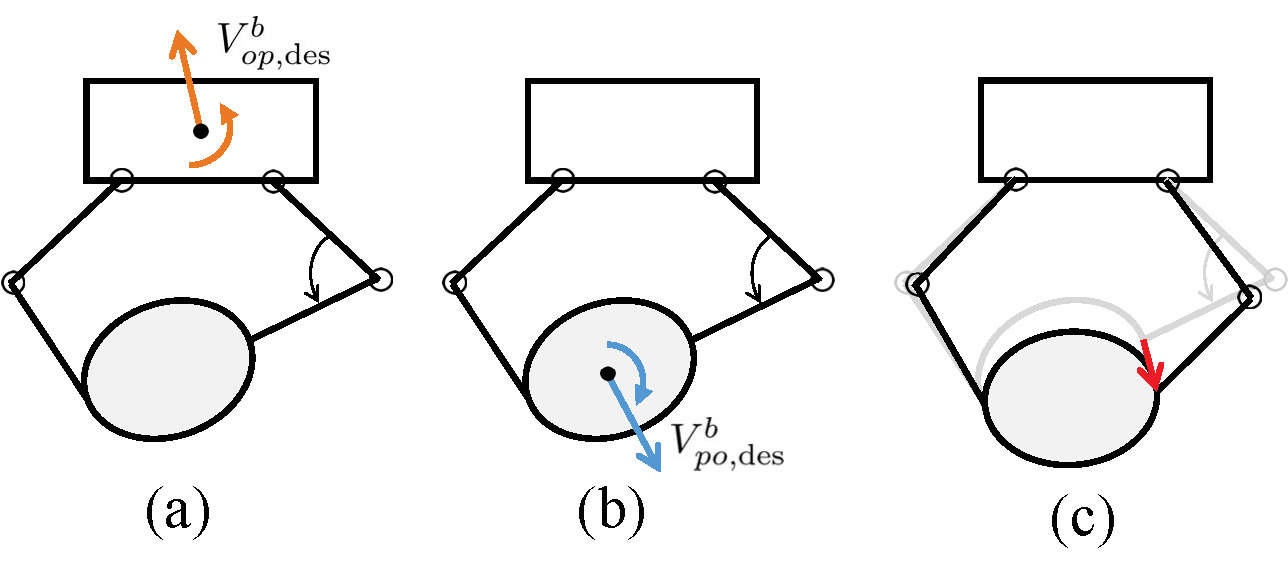}
			\caption{Illustration of the PPO algorithm. (a) shows the desired palm motion, (b) shows the equivalent desired motion of the object based on relativity of motion, and (c) shows the contact estimation and tracking. }
			\label{fig:ppo_illustration}
		\end{center}
	\end{figure}

	Finally, the overall PPO algorithm is summarized in Alg.~(\ref{alg:ppo}). 
	The PPO will be terminated once the quality increment is less than $\delta_{p}$, or angle between the surface normal $\boldsymbol{n_\text{des}}$  and the artificial fingertip normal $\boldsymbol{n_f}(\boldsymbol{q_\text{des}})$ is larger than $\gamma$, as shown in Line~(\ref{ppo:t}). 
	A graphical illustration of the PPO algorithm is shown in Fig.~\ref{fig:ppo_illustration}. Figure~\ref{fig:ppo_illustration}(a) shows the desired tangent space motion of the palm w.r.t. the object (orange arrow lines), and Fig.~\ref{fig:ppo_illustration}(b) shows the equivalent tangent space motion of the object w.r.t. the palm (blue arrow lines) corresponding to Line~(\ref{ppo:object_velocity}) based on the relativity of motion. Fig.~\ref{fig:ppo_illustration}(c) shows the computing and tracking of the desired contacts (red arrow line) corresponding to Line~(\ref{ppo:reference}-\ref{ppo:desire1}).
	\begin{algorithm}
		\caption{Palm Pose Optimization (PPO)}\label{alg:ppo}
		\begin{algorithmic}[1]
			\State \textbf{Input: } Static contact $\boldsymbol{c_0}$ on object, initial joint $\boldsymbol{q}$\label{ppo:input}
			\While {$it_{ppo}\texttt{++} < M$}
			\State Compute object $V_{po,\text{des}}^b, g_{po, \text{des}}$ by~(\ref{eq:ppo_solve}) and~(\ref{eq:ppo_object_update}) \label{ppo:object_velocity}
			\State Compute desired contact $\boldsymbol{c_{\text{des}}^p}, \boldsymbol{v_{c,\text{des}}^p}$ by: \label{ppo:ref_contact}\par
			\hskip\algorithmicindent $c_{\text{des},i}^p \leftarrow R_{po}c_i + t_{po} $ \par
			\hskip\algorithmicindent $v_{c_i,\text{des}}^p \leftarrow R_{po}(v_{po}^b - c_i\times\omega_{po}^b) $
			\label{ppo:reference}
			\State Track desired contact by~(\ref{eq:ppo_track}) \label{ppo:track}
			\State Compute desired joint $\boldsymbol{q_{\text{des}}}$ by $ {\boldsymbol{q_{\text{des}}}} \leftarrow {\boldsymbol{\dot{q}_{\text{des}}}} T_s + \boldsymbol{q}$  \label{ppo:desire1} 
			
			\State Test termination $\Delta Q_\text{des} = Q(\boldsymbol{c_0}, \boldsymbol{q_\text{des}}) - Q(\boldsymbol{c_0},\boldsymbol{q}) $ \par
			\hskip\algorithmicindent$stop = (\Delta Q_\text{des} \leq \delta_{p}) \ \|\  (|\boldsymbol{n_\text{des}}^T\boldsymbol{n_f}(\boldsymbol{q_\text{des}})| \geq \gamma)$ \par
			\hskip\algorithmicindent $collide = col\_detect(\boldsymbol{q_\text{des}},g_{po,\text{des}})$
			\label{ppo:t}
			\If{$stop\  \| \  collide $}  \textbf{ break} \label{ppo:t1}
			
			\EndIf  \label{ppo:t2}
			\State Set state $\boldsymbol{q}\leftarrow \boldsymbol{q_\text{des}}, \  g_{po} \leftarrow g_{po,\text{des}}$ \label{ppo:t3}
			\EndWhile
		\end{algorithmic}
	\end{algorithm}

	\subsection{Iterative CPO-PPO}
	The proposed finger splitting decouples the optimization of palm from contacts, thus is capable to accelerate the computation and implement in real-time applications. 
	The CPO and PPO are iteratively optimized to achieve the finger splitting, as shown in Alg.~(\ref{alg:dual}). 
	\begin{algorithm}
		\caption{Iterative CPO-PPO}\label{alg:dual}
		\begin{algorithmic}[1]
			\State \textbf{Input: } Parallel grasp $\{c_1, c_2, v_{ap}\}$, object mesh $\partial O$\label{dual:input}
			\State \textbf{Init:} Initialize state $\{\boldsymbol{c},\boldsymbol{q},R,t\} \leftarrow Map(c_1,c_2, v_{ap},\partial O)$\label{dual:init}\par
			\hskip\algorithmicindent $it_{cpo} = 0; \ it_{ppo} = 0$
			\While {$True$}
			\State Optimize contacts by Algorithm~(\ref{alg:cpo}):\label{dual:cpo}\par 
			\hskip\algorithmicindent$\{\boldsymbol{c}, \boldsymbol{q}, it_{cpo}\} \leftarrow CPO(R,t,\boldsymbol{c},\boldsymbol{q})$
			\State Optimize palm pose by Algorithm~(\ref{alg:ppo}):\label{dual:ppo}\par 
			\hskip\algorithmicindent$\{(R,t)^{-1},\boldsymbol{q}, it_{ppo}\} \leftarrow PPO(\boldsymbol{c}, \boldsymbol{q}) $
			\If{$it_{cpo} < m$ \textbf{and} $it_{ppo}<m$}\label{dual:t1} \textbf{break}
			\EndIf  \label{dual:t2}
			\EndWhile
		\end{algorithmic}
	\end{algorithm}
	
	The iterative CPO-PPO takes a parallel grasp and object mesh as inputs, as shown in Line~(\ref{dual:input}). A simple function $Map$ can be designed based on the structure of the hands to convert the parallel grasp parameterized by contacts $c_1, c_2$ and approach vector $v_{ap}$ into a grasp for the multi-fingered hand parameterized by $\{\boldsymbol{c},\boldsymbol{q},R,t\}$, as shown in Line~(\ref{dual:init}). The mapped initial grasp is adjusted by the proposed CPO/PPO. 
	The CPO and PPO are optimized sequentially in the loop, as shown in Line~(\ref{dual:cpo}-\ref{dual:ppo}). 
	The iterative CPO-PPO will be terminated if both the CPO and PPO are close to convergence or reaching the constraint boundaries,  
	as shown in Line~(\ref{dual:t1}-\ref{dual:t2}). $m \in \mathbb{Z}^+$ denotes an iteration threshold to stop the finger splitting. A graphical illustration of the iterative CPO-PPO is shown in Fig.~\ref{fig:dual_stage}.

	\subsection{Convergence of the Iterative CPO-PPO}
	The convergence of the proposed iterative CPO-PPO is proved in this section using the global convergence theorem~\cite{luenberger1984linear}. 
	Firstly, we show that CPO converges to a local optimum if solved by Alg.~(\ref{alg:cpo}). Based on the global convergence theorem, the convergence of the CPO requires 1) compact domain, 2) existence of a continuous decent function, and 3) closeness of the algorithmic mapping outside of the solution set. The domain $D=\partial O\times SE(3)\times R^{nq}$ is apparently compact. As for the continuous decent function, we use $-Q(x)$ as the decent function, it is continuous and decent at the outside of the solution set. Furthermore, the algorithmic mapping composited by the tangent space searching $\mathcal{T}$ and nonlinear projection $\mathcal{P}$ is closed in the absence of inequality constraints, since $\mathcal{T}$ is continuous and point-to-point, and $\mathcal{P}$ is closed in  $\mathcal{T}(x)$. Therefore, the CPO solved by Alg.~(\ref{alg:cpo}) converges to a local optimum. Similarly, PPO converges to a local optimum if solved by Alg.~(\ref{alg:ppo}). 
	
	Secondly, we prove that the iterative CPO-PPO converges to a local optimum of~(\ref{eq:general_form}). The composite mapping  $\mathcal{A}_\text{dual}=\mathcal{A}_\text{cpo}\circ \mathcal{A}_\text{ppo}$ is closed since $\mathcal{A}_\text{ppo}$ is a continuous point-to-point mapping and $\mathcal{A}_\text{cpo}$ is closed on $\mathcal{A}_\text{ppo}(x)$, where $\mathcal{A}_\text{cpo}$ is Alg.~(\ref{alg:cpo}) and $\mathcal{A}_\text{ppo}$ is Alg.~(\ref{alg:ppo}). 
	Therefore, we conclude that~(\ref{eq:general_form}) solved by Alg.~(\ref{alg:dual}) converges to a local optimum.

	\section{SIMULATIONS}
	\label{res:sim_exp}
\begin{figure}[t]
		\begin{center}
			\includegraphics[width=1.8in]{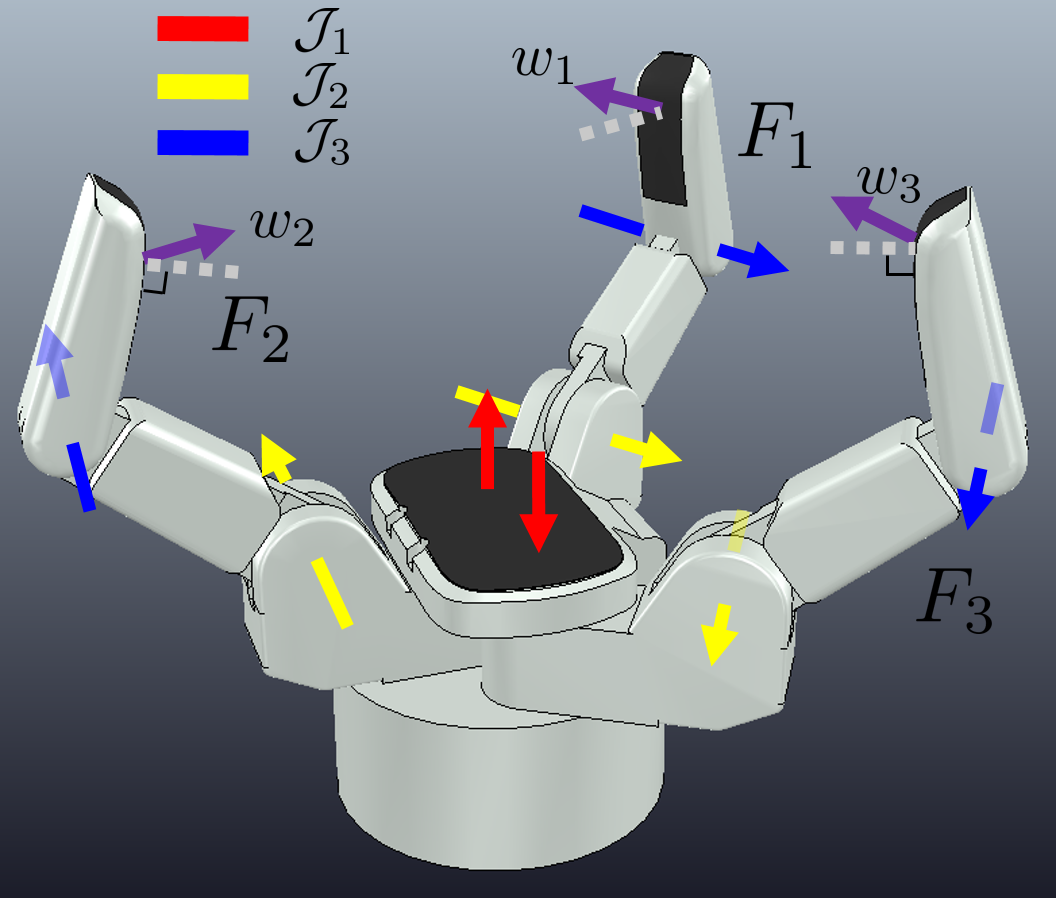}
			\caption{Illustration of the hand structure. }
			\label{fig:hand_model}
		\end{center}
	\end{figure}

	Simulation results are introduced in this section to verify the effectiveness of the iterative CPO-PPO. The simulation video is available at~\cite{youtube}. 
	The grasp planning process is computed in Matlab, visualized in V-REP~\cite{rohmer2013v}, and tested in Mujoco physics engine~\cite{todorov2012mujoco} on a Windows PC with 4.0GHz CPU and 32GB RAM. 
	We use a build-in Barrett hand model but remove joint coupling and enable all eight degree of freedoms (DOFs), as shown in Fig.~\ref{fig:hand_model}. 
	
	\subsection{Parameter Lists}
	Simulation time step $T_s = 0.05$ sec. The weights of grasp quality $w_1 = 1, w_2 = -0.01$ in~(\ref{eq:cpo_nlp}). The maximum iteration $M = 50$ and termination condition $\gamma = 0.6$ in~Alg.~(\ref{alg:cpo}) and~(\ref{alg:ppo}). The minimum iteration bound $m = 2$ in Alg.~(\ref{alg:dual}). 
	The tracking gain $K_{cpo} = K_{ppo}= 2I_9$ and termination condition $\delta_{c} = \delta{p} = 0$. The $\sigma_{cpo}$ and $\sigma_{ppo}$ represent the trust regions of the search to guarantee the accuracy of the approximation of the cost and constraints, and are determined by the smoothness of object surface and kinematics of joints. Empirically $\sigma_{cpo} = 0.15$ and $\sigma_{ppo} = 0.5$. 
	
	\begin{figure*}[htb]
		\begin{center}
			\includegraphics[width=6.5 in]{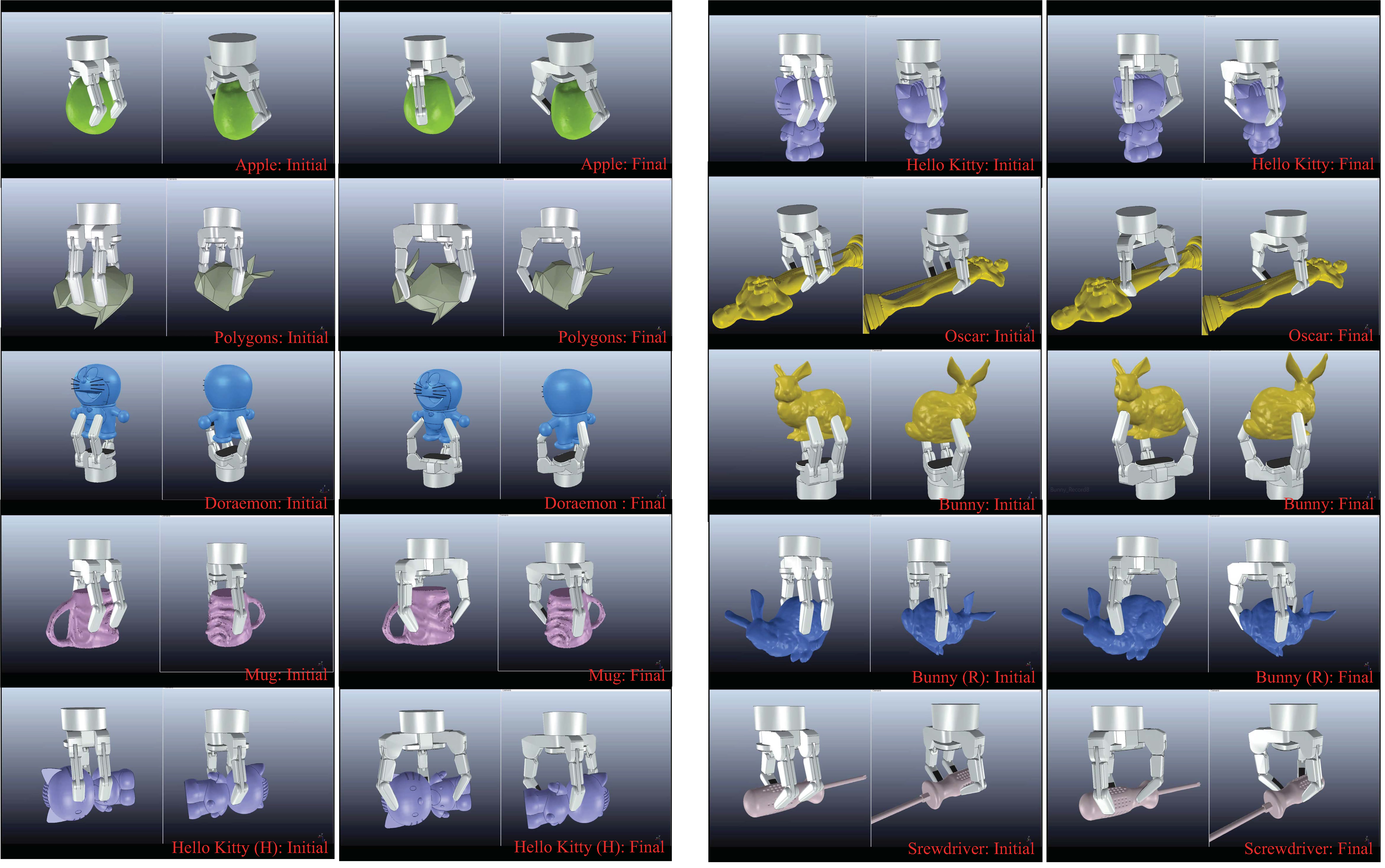}
			\caption{Grasp planning examples for different objects with the multi-fingered hand. }
			\label{fig:result}
		\end{center}
	\end{figure*}
	\subsection{Finger Splitting Results}
	Figure~\ref{fig:result} shows ten grasping examples by the proposed algorithm. 
	The tested objects cover several categories including fruits (apple), toys (Doraemon, Hello Kitty, Oscar), animal (bunny), and tools (mug, screwdriver) with different number of vertices (100 - 150,000). To test the adaptability to different initial conditions, we also test the grasping of the same object with different poses (e.g. Hello Kitty(H) shows the horizontal grasp of Hello Kitty, and bunny(R) denotes grasping of bunny with reverse pose). Each object is first smoothed by~\cite{desbrun1999implicit} then the surface normals are estimated by supervoxel clustering~\cite{Papon13CVPR}.  
	The optimization is able to find all the optimal grasps even the initial parallel grasps are infeasible, as shown in apple and bunny grasps of Fig.~\ref{fig:result}.

	
	Table~\ref{tab:compare} shows the details of grasp synthesis for all ten objects. 
	The 2-5th column shows the number of iterations, the total number of the tangent space searching (and projections), the total optimization time, and the number of vertices of the objects, respectively. 
	In average, for an object with 64K vertices, the iterative CPO-PPO runs for 4.2 outer iterations (84.9/29.2 inner iterations for CPO and PPO, respectively) in total with 0.58 secs in order to find an optimal precision grasp for a three-fingered hand with 8 DOFs. 
	
	
	\begin{table}
		\centering
		\caption{Optimization Details for Grasp Generation}
		\label{tab:compare}
		\begin{tabular}{l|*{4}{l}} 
			Objects    & \#Iters & \#CPO/PPO & Time (ms) & \#Vertices \\
			\hline
			Mug  		& 5 	& 90/22 		& 1257.3	& 185,511\\
			Oscar 		& 2		& 69/3  	& 719.7		& 148,616\\
			Screwdriver & 3		& 40/53 	& 380.6		& 46,721 \\
			Bunny 		& 4 	& 106/28  	& 465.5		& 43,318\\
			Bunny(R) 	& 4 	& 62/44		& 403.7		& 43,318\\
			Doraemon 	& 6		& 125/12		& 484.0		& 42,551\\
			Hello Kitty(H) 	& 5 	& 108/38  	& 467.1		& 29,659\\
			Hello Kitty  	& 7		& 69/56 		& 406.6		& 29,659\\
			Apple 	  	& 1		& 43/2			& 183.0     & 22,487	  \\
			Polygons	&  5	& 96/21	& 1053.1	& 100	 \\
			\hline
			Average   	& 4.20	& 84.9/29.2	& 582.1	& 63,516
		\end{tabular}
	\end{table}

	\begin{table} 
		\centering
		\caption{Average Time Distribution for Grasp Generation}
		\label{tab:time}
		\begin{tabular}{l|*{4}{c}r} 
			Time (ms)       & Tangent search & Projection \& Test & Collision & Total \\
			\hline
			CPO  			& 37.0 	 		& 326.9     & 29.0         & 392.9 \\
			PPO             & 102.9 			& 76.3    & 10.0  & 189.2  \\
			Total           & 139.9 		& 403.2    &  38.9  & 582.1
		\end{tabular}
	\end{table}
	The average time distribution of the optimization for ten grasps is shown in Table~\ref{tab:time}. The tangent space searching (and projection) iterates for 84.9 and 29.2 times for CPO and PPO, respectively. The projection (403.2 ms) takes longer time than tangent space searching (139.9 ms) because of the nearest neighbor computation.
	
\begin{figure}[t]
		\begin{center}
			\includegraphics[width=3.3in]{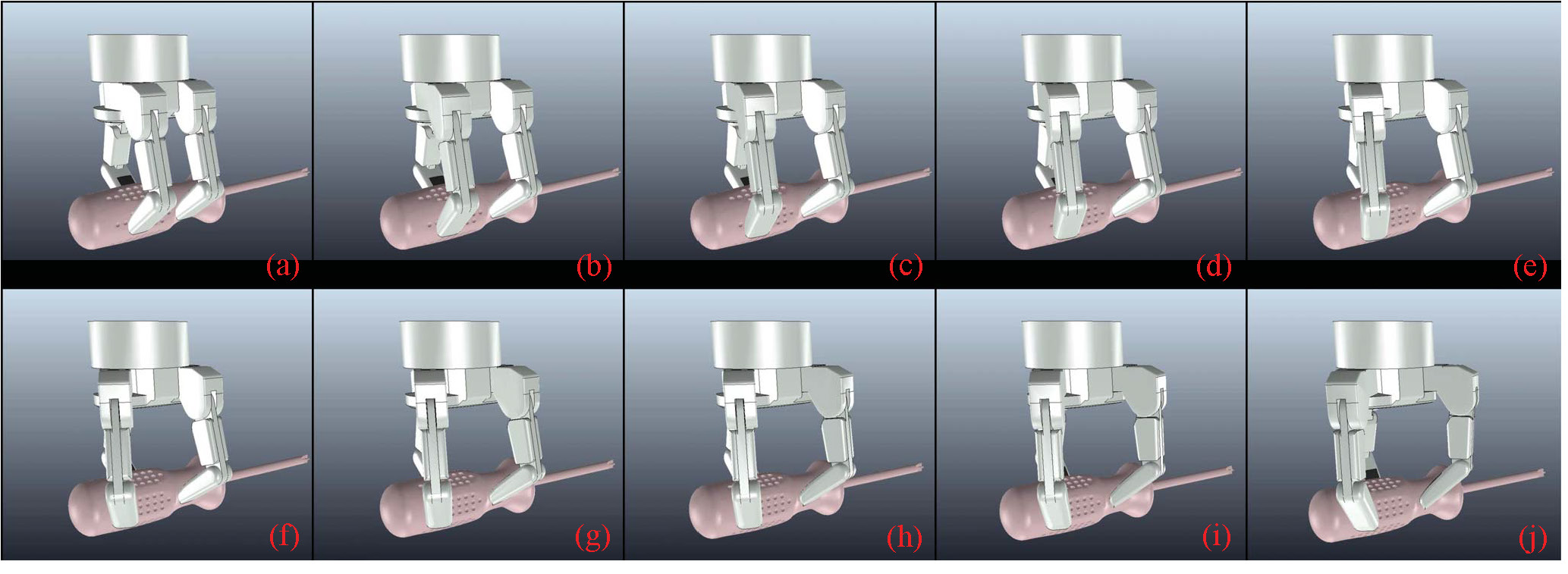}
			\caption{Finger splitting process for screwdriver. The initial and final optimized grasps are in shown in (a) and (j).} 
			\label{fig:animation}
		\end{center}
	\end{figure}

	Figure~\ref{fig:animation} shows the visualization of the iterative CPO-PPO on grasping the screwdriver. The initial grasp generated by a parallel grasp is shown in Fig.~\ref{fig:animation}(a). Then CPO is enabled to solve for optimal contacts by maximizing the overall grasp quality. Due to the successive tangent space searching and projection, the fingers behave as sliding on the object surface, as shown in Fig.~\ref{fig:animation}(b)-(d). The CPO stops when reaching the termination conditions in Alg.~\ref{alg:cpo}. Then PPO is enabled and the pose of the palm is optimized. In practice, the object pose is optimized due to the relativity of motion, as shown in Fig.~\ref{fig:animation}(e)-(g). The PPO and CPO can iterate for several times before converging or reaching the constraint boundaries (e.g. collision or $|\boldsymbol{n_\text{des}}^T\boldsymbol{n_f}(\boldsymbol{q_\text{des}})| \geq \gamma$). The corresponding video is available in~\cite{youtube}. 
	
\begin{figure}[ht]
		\begin{center}
			\includegraphics[width=3in]{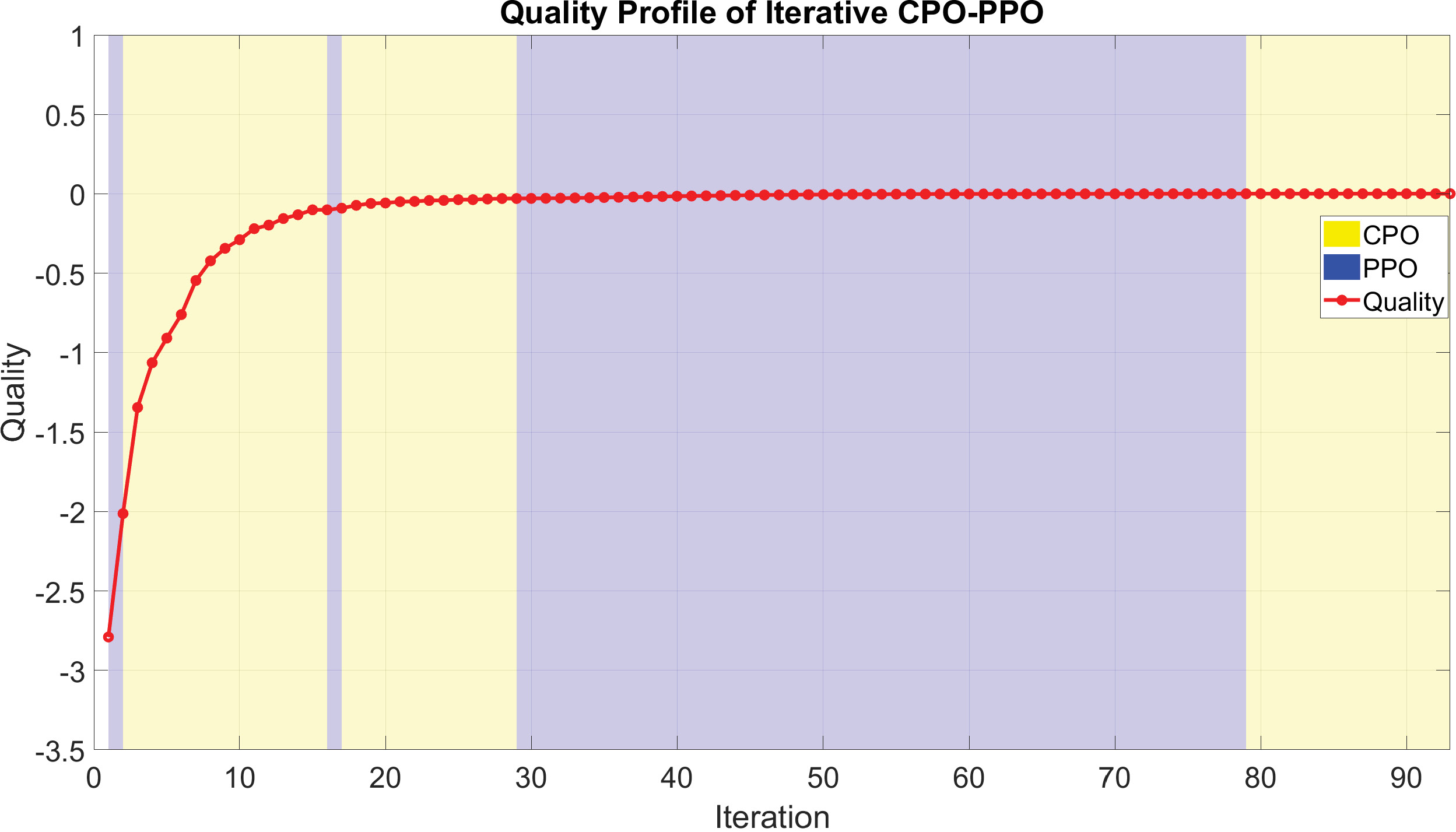}
			\caption{Quality improvement during iterative CPO-PPO for screwdriver. The active CPO and PPO iterations are shown by yellow and blue shaded regions, and the quality profile is shown by red solid curve.}
			\label{fig:quality_profile}
		\end{center}
	\end{figure}

	Figure~\ref{fig:quality_profile} shows the quality profile of the grasp planning on the same object. The active regions for CPO and PPO  are shown by yellow and blue shaded areas, and the quality profile is shown by red solid curve. 
	The algorithm starts from a low-quality parallel grasp and optimizes for joints, contacts and palm by running the iterative CPO-PPO algorithm. The PPO is enabled and terminated in the first iteration, since the $J_1$ is on the joint limit and there is insufficient joint space to improve the overall quality. The CPO is enabled from the second iteration and the contacts are searched by optimizing the overall grasp quality. The CPO runs for 28 iterations, after which PPO takes over and optimizes for object motion. The CPO and PPO iterate until convergence.
	
\begin{figure}[ht]
		\begin{center}
			\includegraphics[width=3.4in]{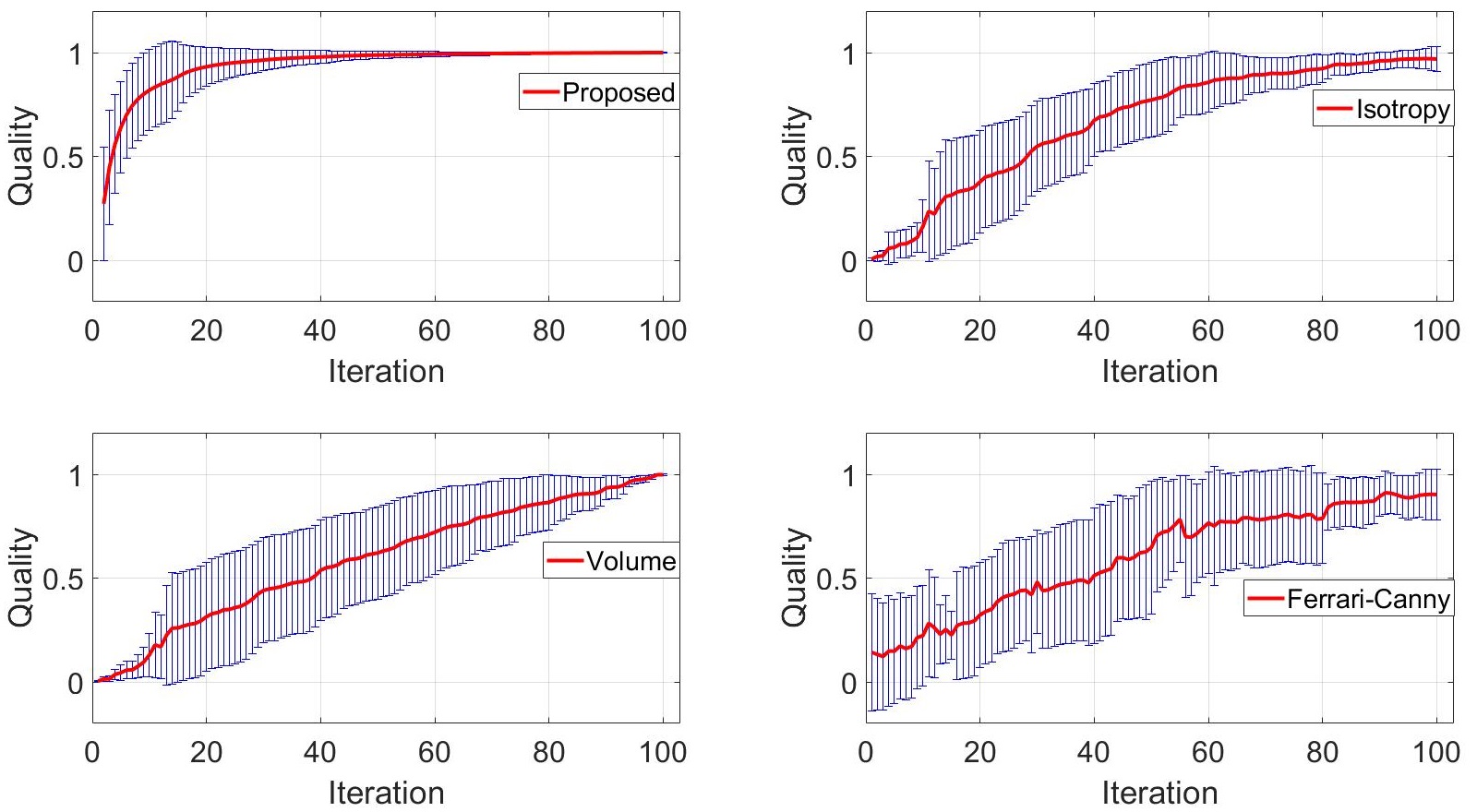}
			\caption{Normalized quality measurements by different metrics during finger splitting. (a) shows the optimization result of the proposed quality metric. (b)(c)(d) are quality measurements of the finger splitting process using grasp isotropy, wrench volume and Ferrari-Canny metrics. }
			\label{fig:qualities_plot}
		\end{center}
	\end{figure}

	Figure~\ref{fig:qualities_plot} shows the measurements of the finger splitting by several commonly adopted quality metrics. The measurements for ten grasps are resized to be the same length. The mean and standard deviation (SD) of the measurements for ten simulated grasps are represented by red sold lines and blue vertical bars, respectively.
	Figure~\ref{fig:qualities_plot}(a) shows the profile of the quality metric optimized in this paper. 
	The average quality is increased monotonically during the finger splitting process. 
	Figure~\ref{fig:qualities_plot}(b) shows the profile of the grasp isotropy index $Q_{iso} = \sigma_{\text{min}}(G)/\sigma_{\text{max}}(G)$~\cite{kim2001optimal}. 
	Figure~\ref{fig:qualities_plot}(c) shows the profile of the wrench space volume metric $Q_{vol} = \sqrt{\text{det}(GG^T)}$~\cite{li1988task}. Figure~\ref{fig:qualities_plot}(d) shows the profile of the Ferrari-Canny metric~\cite{ferrari1992planning}.  
	All the qualities are normalized in order to display the trend during grasp planning of different objects.
	Figure~\ref{fig:qualities_plot}(b)(c)(d) indicate that the finger splitting solved by the iterative CPO-PPO improves grasping performance for all tested grasp metrics. 
	
	\subsection{Simulation Verification}
\begin{figure}[t]
		\begin{center}
			\includegraphics[width=3.22in]{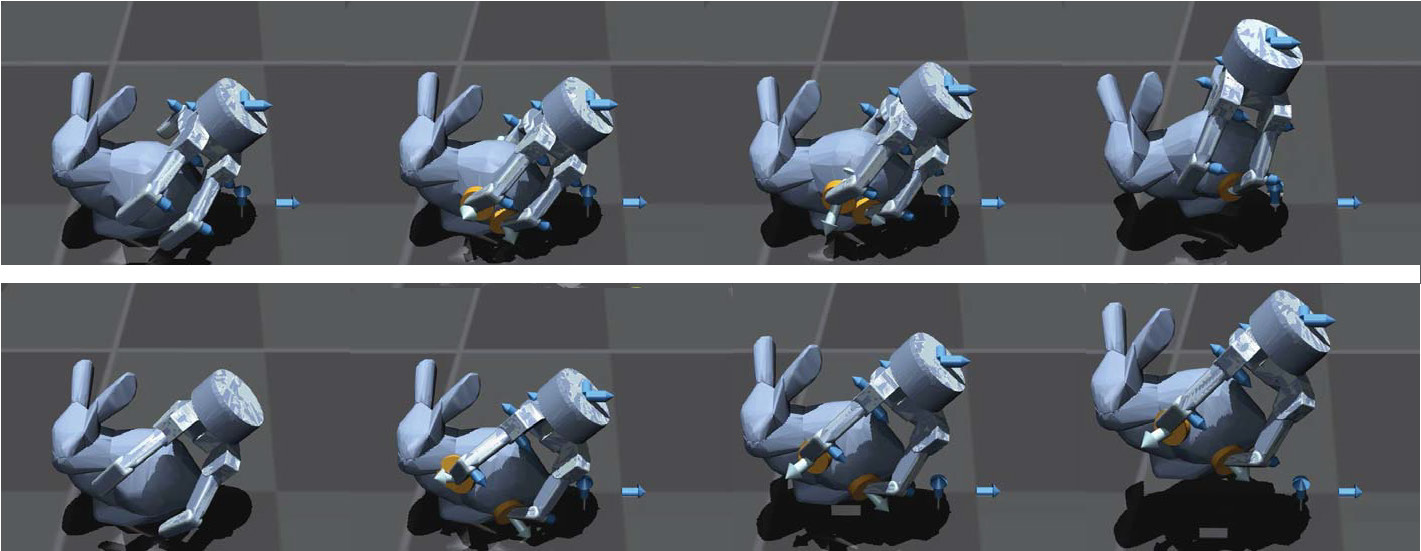}
			\caption{Comparison of the initial optimal parallel grasp (Top) and the finger splitting result (Bottom) in Mujoco. Without finger splitting, the hand cannot grasp object stably due to the gravity and acceleration during lifting.} 
			\label{fig:sim_mujoco}
		\end{center}
	\end{figure}

	The quality improvement of the grasp after finger splitting is verified by the Mujoco physics engine. Given the desired palm pose and contact positions, the fingers are controlled using the virtual frame method~\cite{hang2016hierarchical}. To simulate the nonconvex object for contact and collision, the object is decomposed into convex shapes by v-hacd.
	Figure~\ref{fig:sim_mujoco} shows the simulation results of the initial optimal parallel grasp and the finger splitting result.  The initial parallel grasp cannot lift the object successfully due to the gravitational force and acceleration as shown in Fig.~\ref{fig:sim_mujoco} (Top), while the optimized grasp by finger splitting is able to hold the object steadily during lifting as shown in Fig.~\ref{fig:sim_mujoco} (Bottom). The contact status and magnitude of the contact force are represented by yellow cylinder and gray arrow, respectively. 
	
	\subsection{Comparisons}
	The efficiency of the grasp planning by finger splitting is compared with the methods in~\cite{li2016dexterous} and~\cite{Hang2014Hierarchical} on bunny object, as shown in Table~\ref{tab:compare_methods}. The object surface in method~\cite{li2016dexterous} has to be fitted analytically (takes 8.36 secs), after which an optimal grasp is searched by AMPL using IPOPT solver (takes 15.3203 secs). The hand reachability is not considered in the algorithm. 
	The HFTS planner uses a hierarchical representation of the object. The preprocessing time for the representation is 0.764 secs for the object with 1002 vertices and 13.41 secs for a point cloud with 20586 vertices. The average time for grasp planning on Bunny object is 16.26 secs~\cite{Hang2014Hierarchical}. In comparison, the proposed method achieves the most efficient computation by initializing parallel grasps with ISF~\cite{Fan2018Grasp} (takes 0.15 secs), and searching for optimal grasps for multi-fingered hand by finger splitting (takes 0.43 secs). 
	\begin{table} 
		\centering
		\caption{Computation Time (Seconds) for Different Methods}
		\label{tab:compare_methods}
		\begin{tabular}{l|*{4}{c}r} 
			Methods       & Preprocessing & Optimization  & Total Time \\
			\hline
			Li et. al.~\cite{li2016dexterous}  			& 8.36 	  & 15.32          & 23.68 \\
			HFTS~\cite{Hang2014Hierarchical}           & 0.76 - 13.41	  & 16.26  & 17.02 - 29.67 \\
			\textbf{Proposed}           & 0.15	  &  0.43  & \textbf{0.58 }
		\end{tabular}
	\end{table}
	
	\section{CONCLUSIONS AND FUTURE WORKS} 
	\label{sec:conclusion}
	This paper has proposed a finger splitting strategy for grasp planning with multi-fingered hands by transferring the knowledge from grasp databases of parallel grippers. The splitting  was initialized by the planning result of the parallel gripper, and was optimized continuously by a novel iterative CPO-PPO algorithm. The CPO optimizes for contact points by assuming that the palm is static while PPO optimizes for palm pose by assuming that the contacts on object are static. The CPO and PPO are both solved by consecutive tangent space searching and nonlinear projection. The iterative CPO-PPO algorithm is able to find a local optimal collision-free grasp within one second in average for the objects studied in simulations,  thus is suitable for real-time grasping tasks. 
	
	Current iterative CPO-PPO implementation stops searching if collision is detected or the fingertip normal deviates from the surface normal, thus the optimization can be interrupted before converging. A future work is to add potential fields to push the finger joints away from collision and deviation, as shown in our previous work~\cite{lin2016human}. Future works also include the  
	adaptation of the algorithm to hands with coupled joints and experiments on a Barrett hand using point cloud representation of objects.

	\section{ACKNOWLEDGMENT}
	We gratefully thank reviewers for their comments and suggestions. Thank Prof. Shmuel S. Oren for his advice on the convergence proof. 
	
	\addtolength{\textheight}{-10cm}   
	


	
	
	\bibliographystyle{IEEEtran}
	\bibliography{references}

\end{document}